\UseRawInputEncoding
\documentclass[letterpaper, 10 pt, conference]{ieeeconf}  

\IEEEoverridecommandlockouts                              

\usepackage{algorithm}
\usepackage{algorithmic}
\usepackage{graphicx}
\usepackage{subfigure}
\usepackage{cite}
\makeatletter
\let\NAT@parse\undefined
\makeatother
\usepackage[colorlinks, linkcolor=blue, anchorcolor=green, citecolor=green]{hyperref}
\usepackage{bibentry}

\usepackage{amsmath}

\usepackage{booktabs}
\usepackage{array}
\usepackage{multirow}

\usepackage{authblk}

\title{\LARGE \bf
How to Build a Curb Dataset with LiDAR Data \\for Autonomous Driving
}

\author{Dongfeng Bai, Tongtong Cao, Jingming Guo and Bingbing Liu$^{*}$ 
\thanks{$^{*}$All authors are with Huawei Noah's Ark Lab at the time of writing. \{\tt\small baidongfeng, caotongtong, guojingming, liu.bingbing \}@huawei.com}
}

\begin{document}

\maketitle
\thispagestyle{empty}
\pagestyle{empty}

\begin{abstract}

Curbs are one of the essential elements of urban and highway traffic environments. Robust curb detection provides road structure information for motion planning in an autonomous driving system. Commonly, video cameras and 3D LiDARs are mounted on autonomous vehicles for curb detection. However, camera-based methods suffer from challenging illumination conditions. During the long period of time before wide application of Deep Neural Network (DNN) with point clouds, LiDAR-based curb detection methods are based on hand-crafted features, which suffer from poor detection in some complex scenes. Recently, DNN-based dynamic object detection using LiDAR data has become prevalent, while few works pay attention to curb detection with a DNN approach due to lack of labeled data. A dataset with curb annotations or an efficient curb labeling approach, hence, is of high demand.

In this paper, we present how to build a curb dataset with LiDAR data for autonomous driving highly automatically. Firstly, a Semantic High Definition map (SHD map) in a global coordinate frame is generated by applying both SLAM and semantic segmentation on consecutive LiDAR frames. Next, a Road HD map (RHD map) is generated from the SHD map by removing its dynamic noise caused by road users {\it e.g.} cars. After that, a Curb Instance map (CI map) can be obtained from the filtered RHD map by a series of curb point extraction and growing. Finally, the CI map can be projected back to single frames for direct, highly automatic curb labeling. In order to validate our proposed labeling method, on top of an open public LiDAR semantic dataset SemanticKITTI\cite{Behley2019}, an additional curb dataset is built. We run both semantic segmentation and instance segmentation methods on this built dataset. Experimental results show that the curb annotations have good consistency and accuracy. We released this dataset and it is publicly available at \href{https://download.mindspore.cn} {https://download.mindspore.cn}.

\end{abstract}

\section{INTRODUCTION}

\begin{figure}[t]
	\centering
	\subfigure[SHD map]{\includegraphics[width=2.8cm]{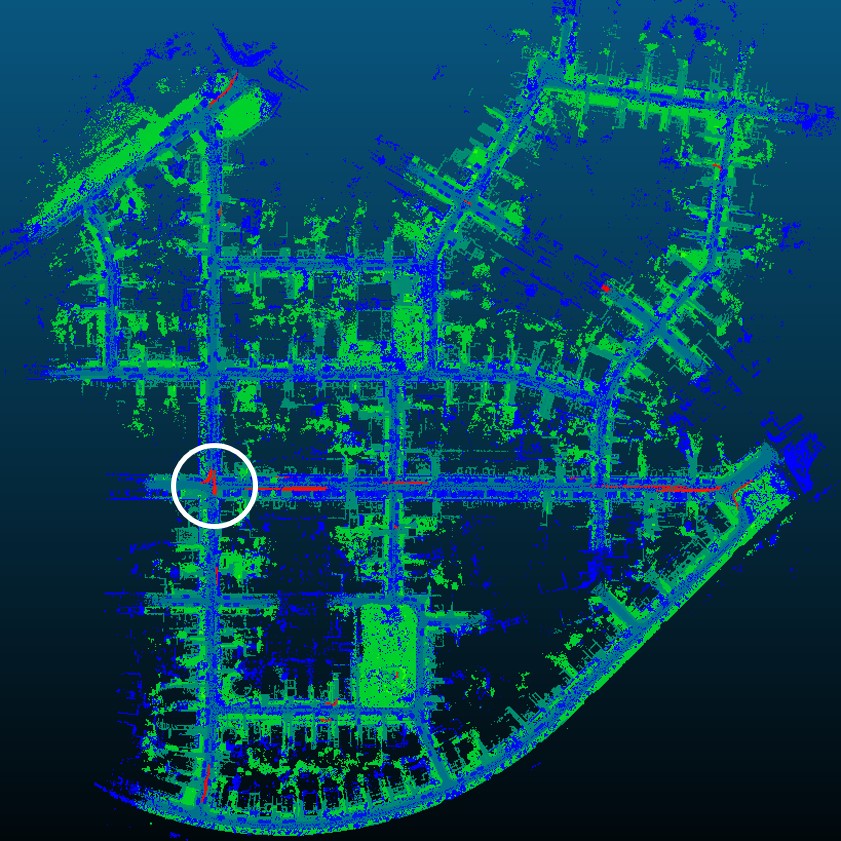}}
	\subfigure[RHD map]{\includegraphics[width=2.8cm]{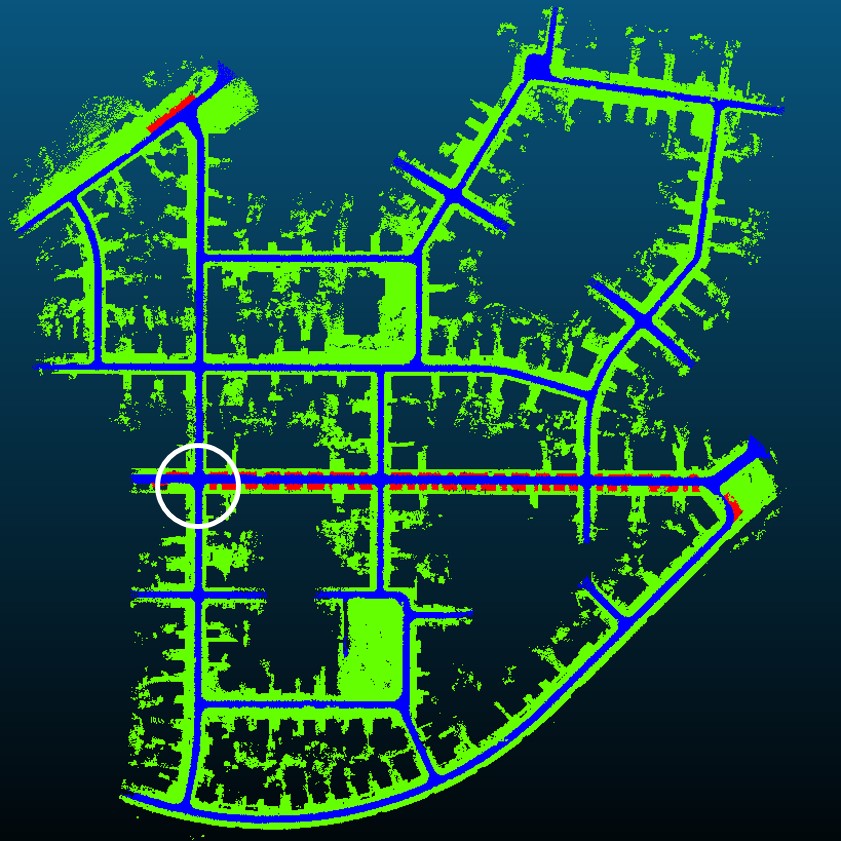}}
	\subfigure[CI map]{\includegraphics[width=2.8cm]{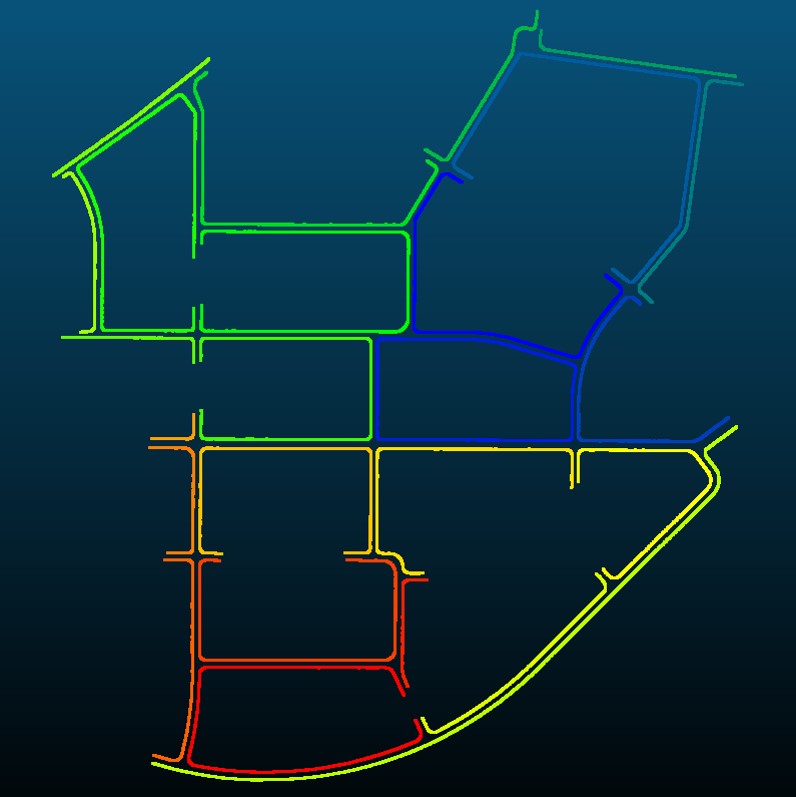}}
	\\
	\centering
	\subfigure[Zoom-in view of the circular region in (a)]{\includegraphics[width=4.26cm]{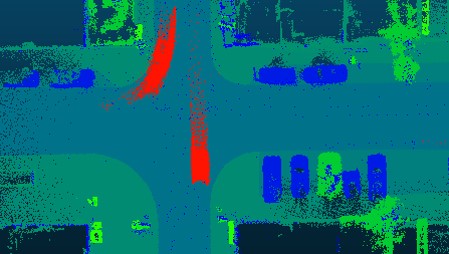}}
	\subfigure[Zoom-in view of the circular region in (b)]{\includegraphics[width=4.26cm]{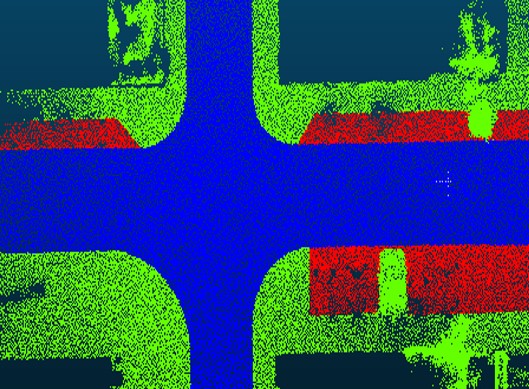}}
	\caption{In the first stage of our labeling method, a CI map is generated with consecutive LiDAR frames. (a) SHD map: generated by a SLAM framework and a semantic segmentation network. (b) RHD map: generated by removing the dynamic noise in (a). Blue pixels are road areas, green pixels are non-road areas, and red ones are parking areas. (c) CI map: output of the first stage, each curb instance is shown in a different color. (d) and (e) are zoom-in views in the circular regions of (a) and (b).}
	\label{fig:pic1}
\end{figure}

In the past few years, autonomous driving has attracted tremendous attention and been developing rapidly. Vehicle-mounted sensors, such as LiDAR, radar and video camera, are extensively utilized in multiple environmental perception tasks ranging from object detection and tracking, semantic segmentation, lane and curb detection for autonomous driving applications. Recently, a various of benchmark datasets have been proposed to satisfy demands of algorithm evaluation and testing. For instance, \cite{Geiger2012, Caesar2020, Sun2020, Wang2019, Cordts2015} collected a large amount of camera and LiDAR data for object detection, tracking and segmentation, and \cite{Tusimple2017}, \cite{Xingang2018} were released for lane line detection based on video camera data. While there are relatively few public benchmarks or datasets available for the challenge of LiDAR-based curb detection, which plays a critical role in road environmental perception. We aim to address this gap and concentrate on how to build such a curb dataset efficiently.

Existing curb labeling methods ({\it e.g.} \cite{Chen2015, Zhang2018, Liang2019}) are mostly based on manual ways. In \cite{Chen2015}, Chen {\it et al.} built a curb dataset containing 2,934 LiDAR scans in various urban scenes and 566 scans in the dataset were labeled manually. Zhang {\it et al.} \cite{Zhang2018} collected about 200 scans in five different scenarios and manually labeled the curbs in each frame. Recently, \cite{Younghwa2021} built and released a curb dataset consisting of about 5,200 scans with BEV labels and encoded images.

Labeling curbs manually are inefficient, costly and error-prone, especially in LiDAR point clouds. Furthermore, due to sparsity of faraway point clouds and blocking by road users, labeling curbs in single LiDAR frames often suffers from partial observations, which makes it provide less useful information in the training of DNN-based curb detection method.

In this paper, we propose an efficient two-stage curb labeling method with LiDAR data. Benefiting from multi-frame consecutive LiDAR data and a CI map, both visible and occluded curbs are labeled simultaneously. The contributions of this paper can be summarized as follows:

\begin{itemize}
	
\item We propose an efficient two-stage curb labeling method which can label LiDAR data with point-wise and instance-wise annotations.
\item We present an annotated curb dataset of LiDAR sequences based on \cite{Behley2019}.
\item We perform curb instance segmentation and semantic segmentation on the labeled dataset and the curb annotations are validated.

\end{itemize}

\section{RELATED WORKS}

\subsection{Road Map Generation}

Curbs and lane lines act as two essential components for road map generation on structured roads (urban roads and highways). Benefiting from high contrast color of lane markers relative to road surface, most road map generation methods rely on lane lines detection using video cameras. \cite{Jeong2017} proposed a Road-SLAM algorithm for road markings mapping and localization. In \cite{Jang2018}, Jang {\it et al.} proposed an automatic HD map generating algorithm with a monocular camera. Qin {\it et al.} \cite{Qin2021} proposed a light-weight mapping and localization solution, which consists of on-vehicle mapping, on-cloud mapping and user-end localization. In contrast, curbs are irregular in color information, but exhibit robust consistency in spatial distribution, distinct from backgrounds. Inspired by this, some methods take advantage of LiDAR's 3D point clouds to detect curbs and build curb maps. He {\it et al.} \cite{He2018} proposed a vector-based road structure mapping method using multi-beam LiDAR and used polyline as primary mapping element. In \cite{Wang2017}, a robust road shape model was proposed and Gaussian process (GP) was employed to generate smooth curves. \cite{Darms2010} presented two approaches for estimating road boundary map by using a radar sensor and a video camera.

\subsection{Curb Detection}


LiDAR-based curb detection methods can be divided into two categories: traditional methods and DNN-based methods. Generally, the traditional methods use hand-crafted features to extract candidate points, which are subsequently clustered and fitted to get parameterized curb results. Due to convenient deployment on computing platforms, most LiDAR-based curb detection methods applied in autonomous driving systems are still traditional ones ({\it e.g.\,}\cite{Chen2015},\cite{Hata2014},\cite{Zhang2018}). However, traditional methods fail in some complex scenarios, such as cross-roads, roundabouts and lower urban curbs (For instance, the height difference is less than $0.1m$ above road surface). Moreover, the hand-crafted features contain a large number of hyperparameters and cannot adapt to different scenarios. DNN-based methods are promising to overcome these constraints, but only several works ({\it e.g.\,}\cite{Suleymanov2019},\cite{Younghwa2021}) were published, due to lack of datasets with LiDAR data for curb detection.

\begin{figure}[htb]
	\centering
	\subfigure[Single-frame]{\includegraphics[width=2.8cm]{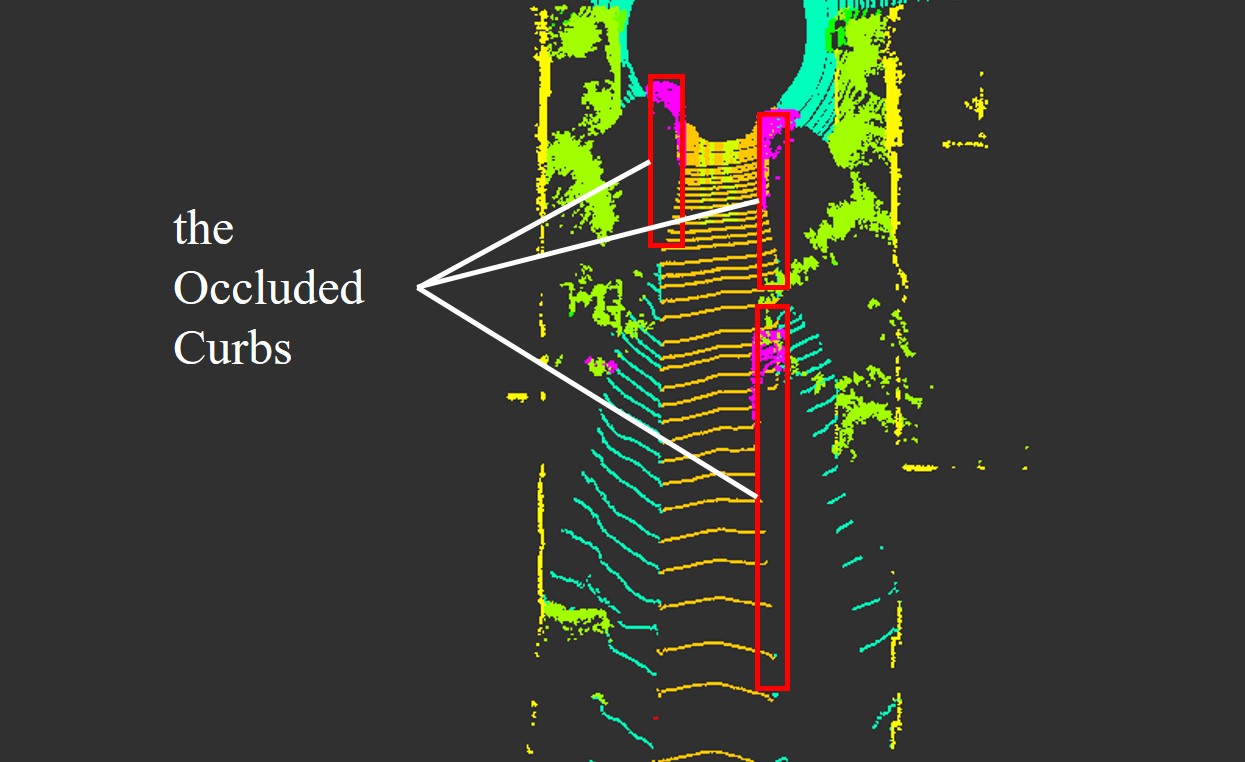}}
	\subfigure[Multi-frame]{\includegraphics[width=2.8cm]{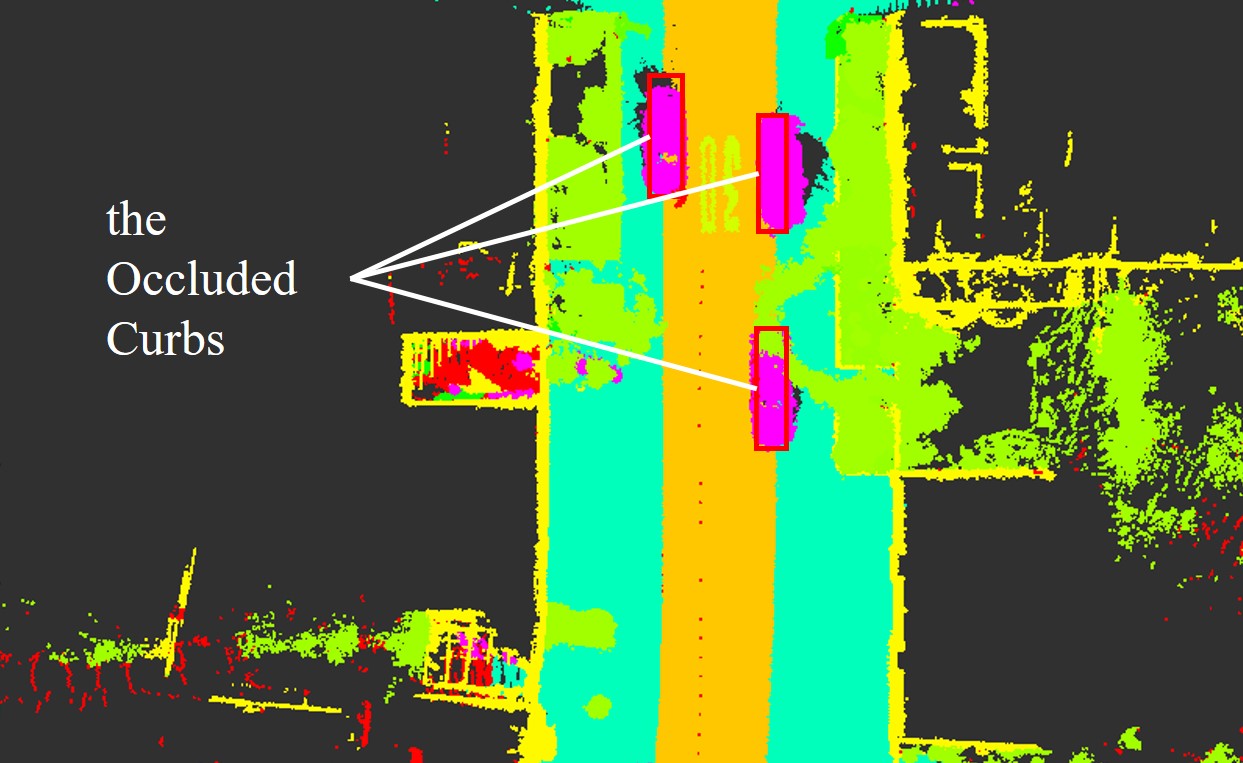}}
	\subfigure[SHD map]{\includegraphics[width=2.8cm]{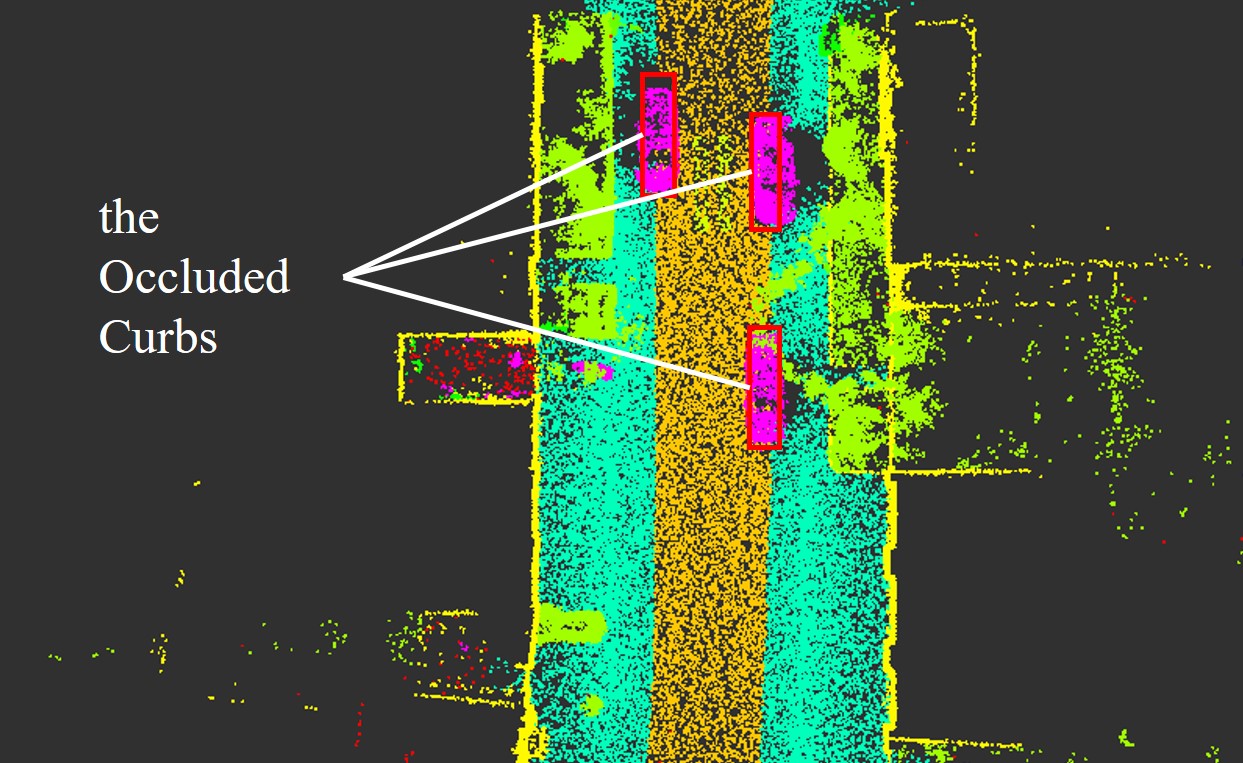}}
	
	\caption{An example of occluded curbs. The occluded curbs are marked with red rectangles. (a) Single-frame LiDAR data is sparse to label the occluded curbs. (b) After superimposing multi-frame data, curbs are more complete, and blind areas are smaller. (c) Blind areas in the corresponding SHD map are similar to (b).}
	\label{fig:pic2}
	\vspace{-0.8em}
\end{figure}

\begin{figure*}[t]
	\centering
	\includegraphics[width=17.8cm]{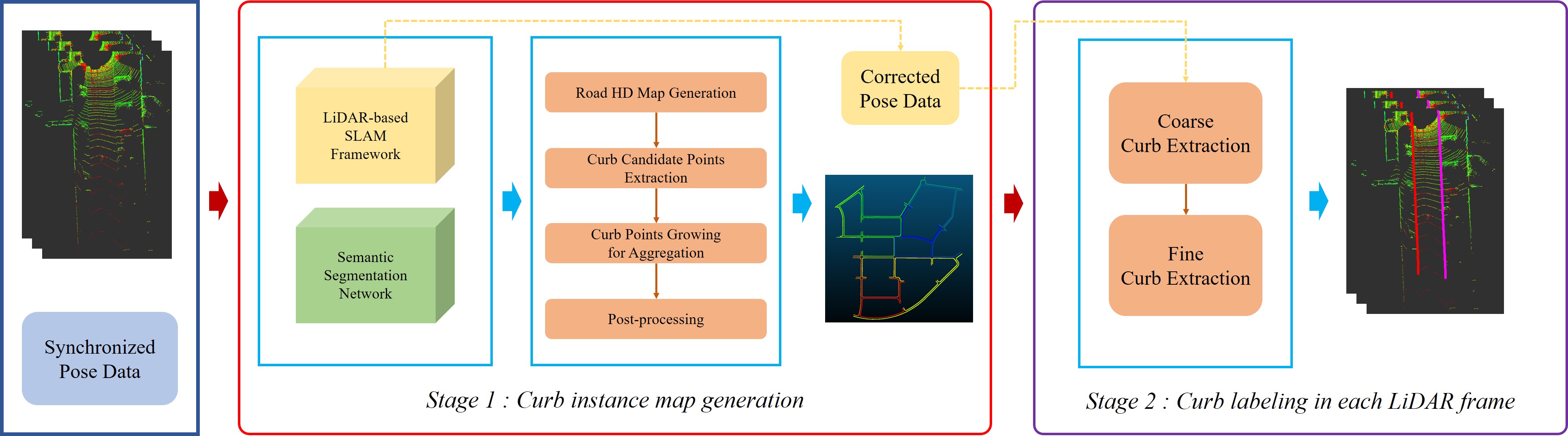}
	\caption{Overview of our proposed two-stage curb labeling method. The input data includes a LiDAR data sequence and synchronized pose data. SHD map is generated by both a LiDAR-based SLAM framework and a semantic segmentation network. A CI map is then generated by 4 steps in Stage 1 and subsequently projected back to single frames for curb labeling in Stage 2. The synchronized pose data is used and corrected in the SLAM framework, and the corrected pose data is utilized in the projection to keep consistency between the CI map and the raw LiDAR data. Coarse curb extraction and fine curb extraction are employed to collect and fine-tune curb annotations respectively.}
	\label{fig:pic3}
\end{figure*}

\section{PROPOSED METHOD}

\subsection{Motivation}

Generally, curbs are the boundaries between a road area and non-road areas such as sidewalks or vegetation. A curb instance is a boundary with continuous spatial distribution along the road, and typically, there are two curb instances on a straight road, and four on cross-roads.

The framework of our two-stage curb labeling method is shown in Fig. \ref{fig:pic3}. A CI map is generated from the LiDAR data sequence and synchronized pose data in the first stage, and then projected back to each LiDAR frame to label curbs in the second stage. The raw LiDAR data and pose data used in the examples and illustrations presented in this paper are from the dataset in \cite{Behley2019}. But our labeling method is adequately general to apply to other similar datasets.

As mentioned above, curbs in single-frame LiDAR point clouds are commonly partially observed for curb labeling in some complex scenarios, due to occlusion or point cloud sparsity. As shown in Fig. \ref{fig:pic2}(a), curbs on the left and right of the road area are occluded by three static cars. Fortunately, curbs are unmovable and smooth, thus after multi-frame point clouds are accumulated by pose data, as shown in Fig. \ref{fig:pic2}(b)-(c), the boundaries between roads and sidewalks become more complete on account of multiple observations from a set of varied perspectives. Therefore, labeling curbs in multi-frame LiDAR data or in a SHD map can be promisingly achieve more accurate results and also have the advantage of high efficiency.

\subsection{Stage 1: Curb Instance Map Generation}
In this stage, we generate a global CI map with LiDAR data and synchronized pose data to prepare for the curb labeling for a LiDAR data sequence. It is not advisable to roughly utilize the pose data to superimpose the multi-frame LiDAR data, which will lead to global inconsistency for a data sequence with loop closures. As mentioned in \cite{Behley2019}, streets which are revisited have different heights if multiple scans are superimposed above each other in a simple way.

\subsubsection{RHD Map Generation}In the first stage of our method, a semantic segmentation network\cite{Gerdzhev2020} and a LiDAR-based SLAM system\cite{Pan2021} are employed to build a SHD map as shown in Fig. \ref{fig:pic1}(a). In \cite{Gerdzhev2020}, the novel semantic segmentation technique divides raw point clouds into 20 categories, while the categories of road users such as vehicles, pedestrians, etc., are removed in RHD map since they are irrelevant for extracting curbs, as shown in Fig. \ref{fig:pic1}(b). Then we subdivide the RHD map into several RHD sub-maps in order to parallelize subsequent map processing and labeling.

\begin{figure}[h]
	\centering
	\includegraphics[width=8cm]{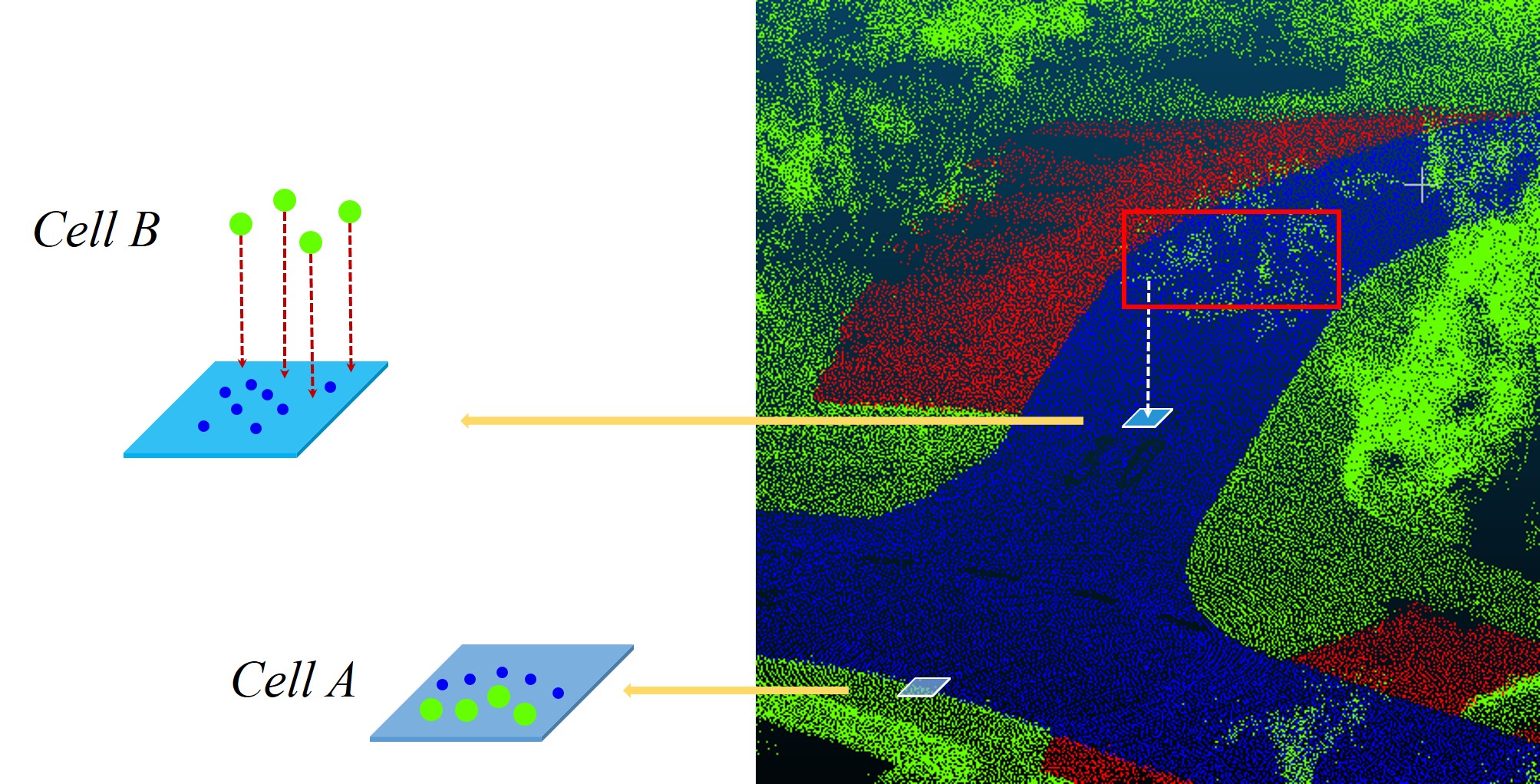}
	\caption{Illustration of how to extract curb candidate points. Cell A belongs to a curb cell, while Cell B belongs to a road cell.}
	\label{fig:pic4}
	\vspace{-0.8em}
\end{figure}

\subsubsection{Curb Candidate Points Extraction}For each RHD sub-map, an empty 2D grid map is applied and the sub-map's point cloud is projected into it for fast extracting of curb candidate points. Cells in the grid map could be divided into 4 categories: road cell, non-road cell, curb cell and unknown cell. A road cell only contains road points. A non-road cell contains non-road points, and there is no point in unknown cells. As for a curb cell, it should satisfy the condition that both road points and non-road ones are contained in the cell. However, point clouds of trunk and vegetation often invade the airspace of road area as shown in Fig. \ref{fig:pic4}. Hence, an extra height condition is induced for curb cells: height distributions of road points and non-road points in each cell must be similar. Only the cells meeting both of the above two conditions are curb cells and the points in these curb cells are curb candidate points.

\begin{figure}[h]
	\centering
	\includegraphics[width=9.2cm]{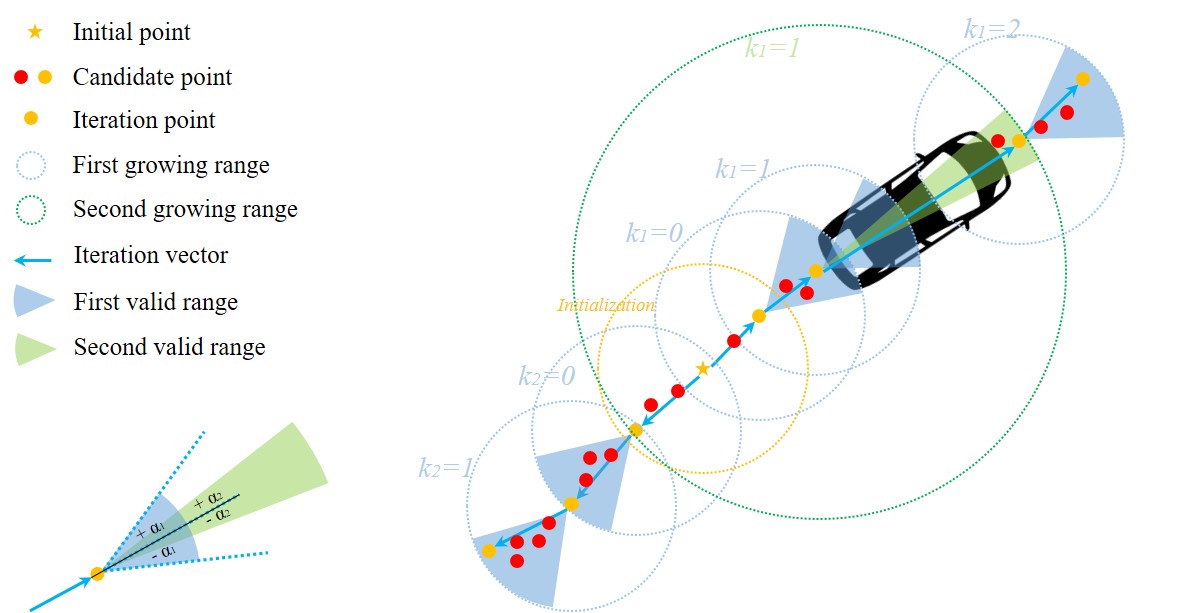}
	\caption{Illustration of Alg. \ref{alg1}. To deal with occluded curbs, we design two growing ranges in function $curbgrow$. The first one is the same with an initial range $R_1$ and with a valid range ($-\alpha_1, +\alpha_1$). The second one is larger but with a narrower valid range ($-\alpha_2, +\alpha_2$). A valid range is a fan-shaped region and its orientation is determined by the corresponding iteration vector.}
	\label{fig:pic5}
	\vspace{-0.8em}
\end{figure}

\subsubsection{Curb Points Growing for Aggregation} Curb candidate points in each RHD sub-map are disordered and a clustering algorithm needs to be implemented to generate curb instances. Since the distribution of curb candidate points is narrow and uneven, classical clustering algorithms, such as K-means\cite{MacQueen1967} and DBSCAN (Density-Based Spatial Clustering of Applications with Noise)\cite{Ester1996}, often mistakenly divide a curb instance into several clusters or determines some sparse curb candidate points as noise. Therefore, we design an algorithm based on K-nearest neighbors (KNN) algorithm\cite{Altman1992} for simultaneous clustering and sorting curb candidate points. The detailed procedure is shown in Alg. \ref{alg1}. Function $curbgrow$ is a dual growth strategy and is designed in iterations of curb growing. As shown in Fig. \ref{fig:pic5}, if there is no curb candidate point within the first valid range due to obstruction of a parked car or other obstacles, the second growth takes effect, which has a larger growing range but a narrower valid range than the first growth. Once the growing iteration of the $j$th curb completes, point array $C_{res2}$ is flipped and combined with $C_{res1}$ and $\left\{p_{init}\right\}$ as the final ordered point array of the $j$th curb instance. After the procedure in Alg. \ref{alg1}, all curb candidate points are clustered into multiple sets of curb points, and the points in each set are ordered.
\begin{algorithm}[t]
	\caption {Curb Candidate Points Clustering}
	\label{alg1}
	\begin{algorithmic}[1]
		\REQUIRE ~~\\
			Curb candidate points set ${\bf P}$ = $\left\{ {p}_{i} \right\}$ in a sub-map
		\ENSURE ~~\\
			Multiple sorted and clustered point sets ${\bf C}$ = $\left\{ {C}_{j} \right\}$
		\STATE Build a k-d tree ${\bf T}$ with the input point set ${\bf P}$
		 
		\STATE Un-queried set ${\bf P}_{todo} \gets {\bf P}$
		\STATE $j \gets 1$
		\WHILE { ${\bf P}_{todo} \neq \emptyset$}
		\STATE Pick an initial point ${p}_{init}$ in ${\bf P}_{todo}$ randomly and query the neighbors ${C}_{nei}$ of ${p}_{init}$ within a circular range $R_1$ in ${\bf T}$. The point number of ${C}_{nei}$ is ${N}_{nei}$. 
		\STATE ${\bf P}_{todo} \gets {\bf P}_{todo} - \left( C_{nei} \cup \left\{p_{init}\right\} \right)$
		\IF { ${N}_{nei} >= \psi$ }
		\STATE Subdivide $ C_{nei} $ into $ C_1^0 $ and  $C_2^0$ by azimuths
		\STATE Sort $C_1^0$, $C_2^0$ in ascending order of 2D distance from ${p}_{init}$ as $\tilde{C}_1^0$, $\tilde{C}_2^0$ 
	    \STATE $p_1^0$, $p_2^0$ are the iteration (farthest) points in $\tilde{C}_1^0$, $\tilde{C}_2^0$ 
	    \STATE Iteration vectors: $d_1^0 \gets \overrightarrow {p_{init} p_1^0}, d_2^0 \gets \overrightarrow {p_{init} p_2^0}$
		\STATE Iteration sets: ${C}_{res1} \gets \tilde{C}_1^0$, ${C}_{res2} \gets \tilde{C}_2^0$
		\STATE Iteration times: ${k}_{1} \gets 0$, ${k}_{2} \gets 0$
		\STATE Iteration flags: ${F}_{1} \gets true$, ${F}_{2} \gets true$
		\WHILE {${F}_1 = true $}
        \STATE $\tilde{C}_1^{k_1+1}, p_1^{k_1+1}, d_1^{k_1+1}, F_1 = $ curbgrow$\left({p}_1^{k_1}, {d}_1^{k_1}\right)$
		\STATE ${\bf P}_{todo} \gets {\bf P}_{todo} - \tilde{C}_1^{k_1+1} $
		\STATE $C_{res1} \gets C_{res1} \cup \tilde{C}_1^{k_1+1}$;
		\STATE ${k_1} \gets {k_1} + 1 $
		\ENDWHILE
		\WHILE {${F}_2 = true $}
		\STATE $\tilde{C}_2^{k_2+1}, p_2^{k_2+1}, d_2^{k_2+1}, F_2 =$ curbgrow$\left({p}_2^{k_2}, {d}_2^{k_2}\right)$
		\STATE ${\bf P}_{todo} \gets {\bf P}_{todo} - \tilde{C}_2^{k_2+1} $
		\STATE $C_{res2} \gets C_{res2} \cup \tilde{C}_2^{k_2+1}$;
		\STATE ${k_2} \gets {k_2} + 1 $
		\ENDWHILE
		\STATE ${C}_{j} \gets C_{res1} \cup \left\{p_{init}\right\} \cup $ flip$ \left(C_{res2}\right) $
		\STATE $j \gets j + 1$
		\ENDIF
		\ENDWHILE
		\RETURN ${\bf C}$ = $\left\{ {C}_{j} \right\}$
	\end{algorithmic}
\end{algorithm}


\subsubsection{Post-processing}
Curb Candidate Points Extraction and Curb Points Growing for Aggregation can be parallelized for the multiple subdivided RHD sub-maps. When a RHD map was subdivided into multiple sub-maps, curbs crossing adjacent sub-maps would be split into several pieces. Therefore, while merging curb points from all sub-maps to one global CI map, it is necessary to integrate the split pieces and re-number all curb point sets. The split curbs to be integrated should meet the following conditions: the distance between two endpoints in adjacent sub-maps is less than $D_{link}$ and the orientation angle between the curbs should be less than $\theta_{link}$.



\subsection{Stage 2: Curb Labeling in Each LiDAR Frame}

\subsubsection{Coarse Curb Extraction}
The CI map is in a global coordinate frame, and it is necessary to transform it to each LiDAR coordinate frame by using the corrected pose data. The transformation is presented in Eq. \ref{eq:a1}.
\begin{equation}
	\begin{aligned}
	{M_{r,j}} & =  \left\{ \parallel P - T_j \parallel  <=  R_2 \right\}, \qquad {P \in M} \\
	{M_{j}} & =  \left[ R_j | T_j  \right]^{-1} * {M_{r,j}}
	\end{aligned}
\label{eq:a1}
\end{equation}
where $M$ is the CI map and $P$ is the curb point in $M$. $R_j$ and $T_j$ are rotation and translation in the $j$th frame in the pose data. ${M_{r,j}}$ is the curbs of $M$ within a circular range $R_2$ around the position of LiDAR and ${M_{j}}$ is the transformed curb annotations corresponding to the $j$th frame of LiDAR data. $R_2$ is slightly larger range than ROI of LiDAR.

If there are $K$ curb instances in ${M_{j}}$ totally, we describe these curb instances as Eq. \ref{eq:a3}.
 \begin{equation}
	\begin{aligned}
		{M_{j}} & =  \left\{ Curb_{j,k}\right\}_{k=1}^{K}
	\end{aligned}
	\label{eq:a3}
\end{equation}

\begin{figure}[htb]
	\centering
	\includegraphics[width=8.4cm]{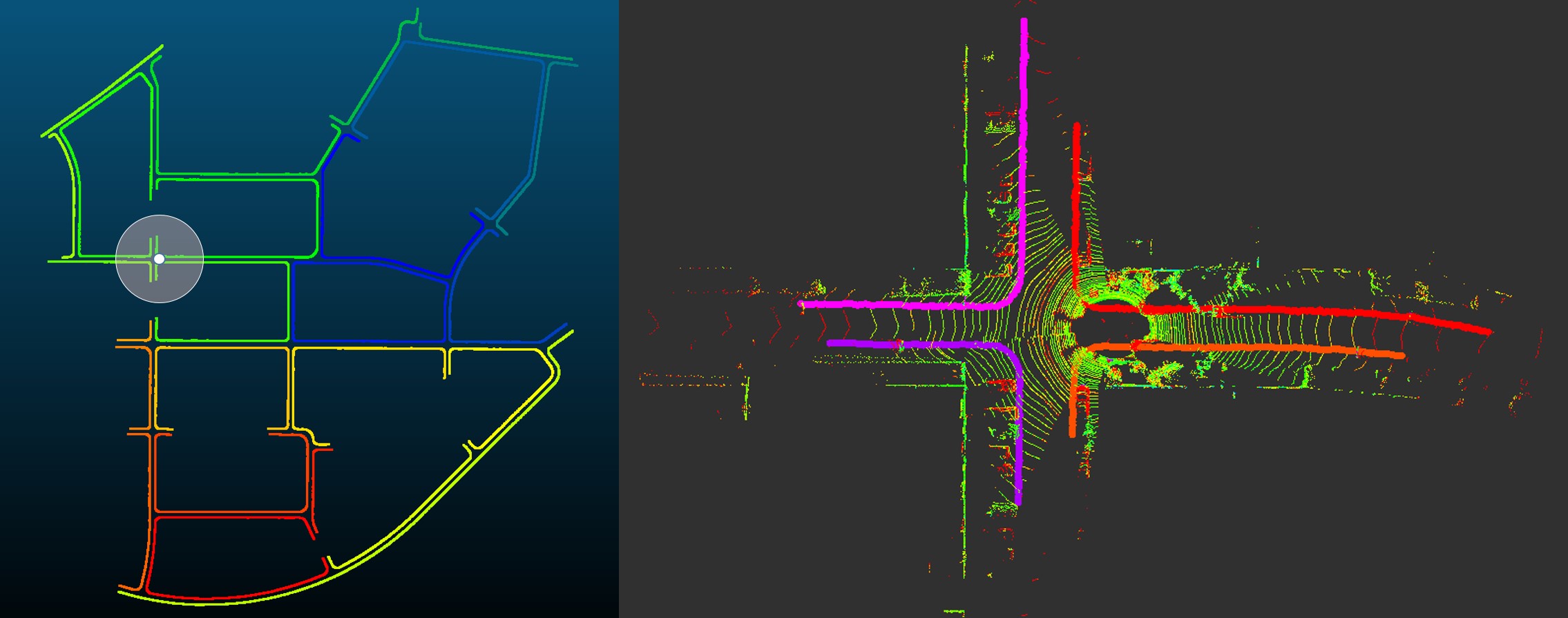}
	\caption{Curb labeling in a single LiDAR frame with a CI map.}
	\label{fig:pic6}
\end{figure}

\subsubsection{Fine Curb Extraction} 
In this step, we determine the lengths and endpoints of curb instances according to the distributions of road points and curb-related (sidewalk or vegetation) points in each single-frame LiDAR data. Eq. \ref{eq:a4} is used to judge whether the curb points in $M_j$ satisfy the condition of fine extraction.
 \begin{equation}
	\begin{aligned}
		\text{FineExtract}({P_{j,k,i}}) & = \text{Count}(P_{j,k,i}, \left\{{P_{road,j}}\right\}, R_3) \\
									    & + \kappa * \text{Count}(P_{j,k,i}, \left\{{P_{side,j}}\right\}, R_4)
	\end{aligned}
	\label{eq:a4}
\end{equation}
where $P_{j,k,i}$ indicates the $i$th point in $Curb_{j,k}$. $\left\{{P_{road,j}}\right\}$ and $\left\{{P_{side,j}}\right\}$ are point clouds of road category and curb-related category respectively. $R_3$ and $R_4$ are radii of circular range around $P_{j,k,i}$. Function $Count$ calculates the number of points in $\left\{{P_{road,j}}\right\}$ or $\left\{{P_{side,j}}\right\}$ within the corresponding circular range around $P_{j,k,i}$. $\kappa$ is a weighting factor.

Then the following Eq. \ref{eq:a2} is used for fine curb extraction.
 \begin{equation}
 	\left\{
 	\begin{aligned}
 		{Index_{j,k}} & =  \left\{ i \right\},  {\quad if } \text{ FineExtract} \left(P_{j,k,i}\right) > \phi  \\
 		{IdS_{j,k}} & =  min\left( Index_{j,k} \right) \\
 		{IdE_{j,k}} & =  max\left( Index_{j,k} \right) \\
 		{FineCurb_{j,k}} & = \left\{{P_{j,k,i}}\right\}, \quad {{IdS_{j,k}}}<i<{{IdE_{j,k}}}
 	\end{aligned}
 	\right.
 	\label{eq:a2}
 \end{equation}
where ${Index_{j,k}}$ is index set of the curb points satisfying the function $FineExtract$. ${IdS_{j,k}}$ and ${IdE_{j,k}}$ are indexes of fine extraction endpoints of $Curb_{j,k}$. ${FineCurb_{j,k}}$ is the $k$th curb annotation in the $j$th frame after fine extraction. $\phi$ is a score threshold of $FineExtract$.

In Eq. \ref{eq:a2}, in order to keep continuity of curb annotations, we determine the lengths and endpoints of curb annotations by the maximum and minimum indexes in ${Index_{j,k}}$. Furthermore, we utilize segmental spline curves to fit the curb annotations and re-sample them with an equal interval to get the final curb annotations.


\section{EXPERIMENT}
Our curb dataset could support both the semantic segmentation and instance segmentation, and performances of these tasks, in turn, could help confirm the validity of our dataset.
We conduct experiments of curb dataset generation on the SemanticKITTI dataset\cite{Behley2019}, which is a large-scale dataset widely used for semantic segmentation with LiDAR data and consists of 22 sequences, splitting sequences 00-10 as the training set, and 11-21 as the test set. The file format of our curb dataset follows \cite{Behley2019}, and the curbs in sequences 00-10 are labeled by our labeling method, totaling 23,201 frames, 55,013 curb instances, and 23,149,310 curb points.
\vspace{-0.8em}
\begin{figure}[h]
	\centering
	\includegraphics[width=8.4cm]{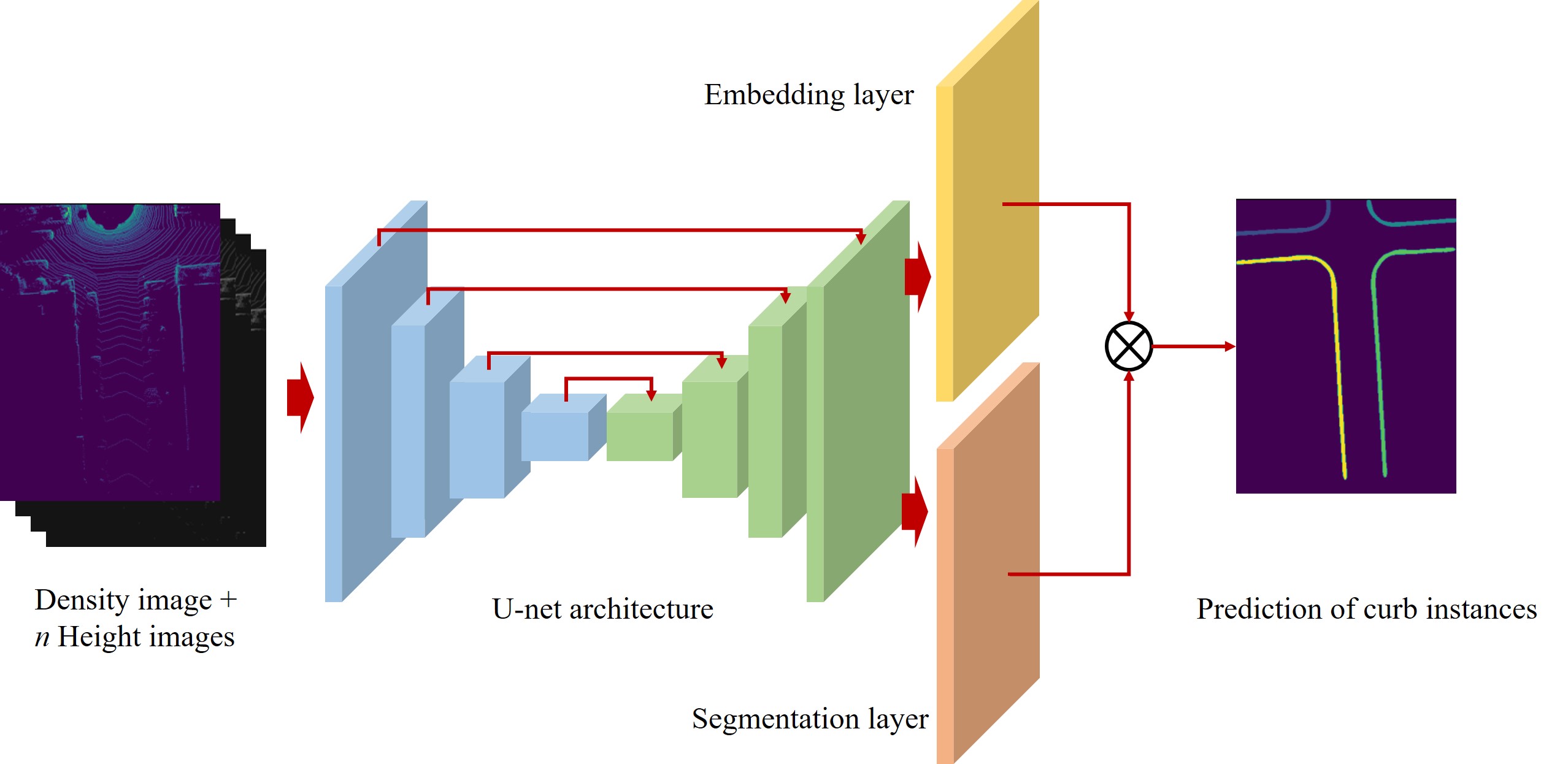}
	\caption{Overview of our curb instance segmentation network.}
	\label{fig:pic9}
	\vspace{-0.8em}
\end{figure}
\subsection{Build a Curb Dataset}

\subsubsection{Stage 1}
In the curb candidate points extraction, the resolution of 2D grid map is set to (0.2$m$ $\times$ 0.2$m$). In each cell, we adopt average height of points indicating the height distribution. Height difference threshold between average heights of road points and non-road points is set to 0.3$m$. In the curb points growing for aggregation, the point number threshold $\psi$ is set to $6$. The first growing range (the initial searching range) is set as $R_1 = 2.6m$ and the first valid angle range is set as $\alpha_1 = 30^\circ$.  The second growing range and the second valid angle range $\alpha_2$ are set to $3R_1$ and $\alpha_1/3$, respectively. 
\begin{figure*}[h]
	\centering
	\includegraphics[width=17.8cm]{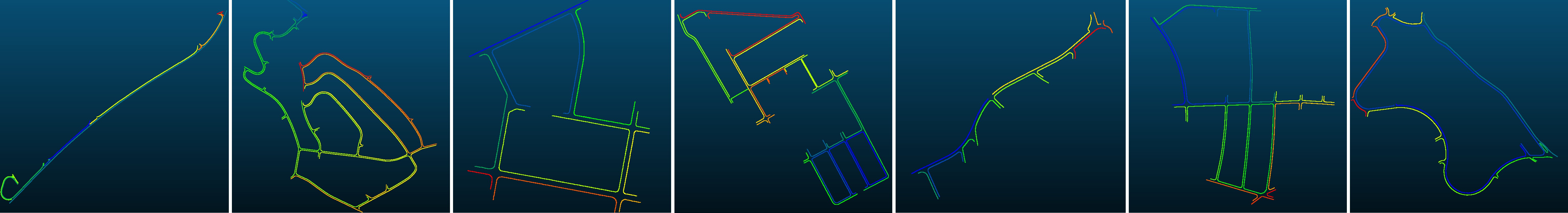}
	\caption{Illustrations of CI maps. Different curb instances are shown in different colors in each CI map.}
	\label{fig:pic7}
\end{figure*}
In the post-processing, $D_{link}$ and $\theta_{link}$ are set to 0.5$m$ and $20^\circ$. After merging the CI sub-maps, we interpolate the points of each curb instance to an interval of $0.1m$. Some CI maps are visualized in Fig .\ref{fig:pic7}.
\subsubsection{Stage 2}
\begin{figure*}[h]
	\centering
	\includegraphics[width=17.8cm]{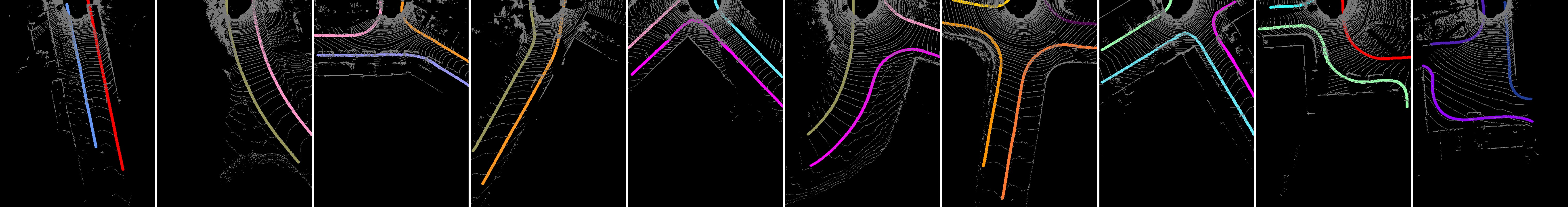}
	\caption{Examples of our curb annotations for curb instance segmentation in single-frame LiDAR data. Dilating kernel pixel size is set to 7.}
	\label{fig:pic8}
\end{figure*}

The coarse extraction range $R_2$ is set to $80m$. The parameters in fine curb extraction are set as: $\kappa = 0.2, R_3 = 3m, R_4 = 5m, \phi = 20$. Some examples of curb annotations with raw point clouds in BEV representation are shown in Fig. \ref{fig:pic8}. 

\subsection{Curb Instance Segmentation}

We formulate curb instance segmentation as a pixel-wise segmentation task and present a curb instance segmentation network. As shown in Fig. \ref{fig:pic9}, U-net\cite{Ronneberger2015} is adapted as the basic architecture. With the preprocessing of LiDAR data and curb labels in \cite{Younghwa2021}, single-frame point cloud is encoded by a density image and multiple sliced height images in BEV representation. Curb annotations are projected into images as labels and the curb pixels are dilated to balance the proportion between positive and negative pixels for training. Furthermore, an embedding decoding layer is utilized for curb instance segmentation referring to the embedding branch in \cite{Neven2018}. After masking the pixels of the embedding layer with the binary segmentation result from the segmentation layer, DBSCAN is used to cluster the embedded points for curb instance extraction.

\begin{figure*}[h]
	\centering
	\includegraphics[width=17.8cm]{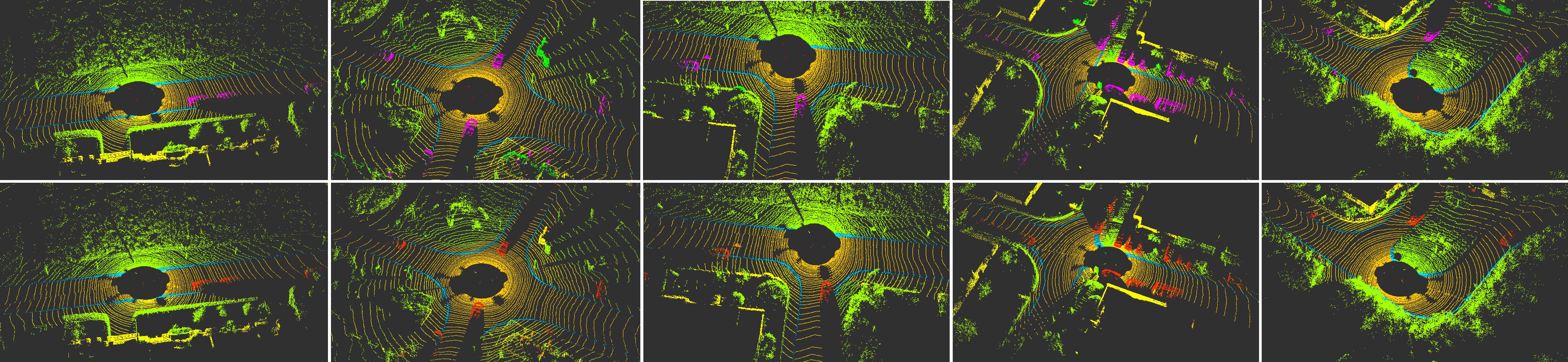}
	\caption{Ground truths and predictions in semantic segmentation with curbs. The first row shows ground truths and the second row shows predictions. Curb points are displayed in indigo blue color.}
	\label{fig:pic10}
	\vspace{-0.46em}
\end{figure*}

We evaluate the performances of curb binary segmentation upon our dataset and the dataset in \cite{Younghwa2021}, \cite{Suleymanov2019}. The basic architecture of binary segmentation in our instance segmentation network shares the same structures with them. Table \ref{tab:1} shows that our curb dataset performs as well as the other manual-labeling dataset. The resolution of BEV representation is set to $0.1m/pixel$. 1 pixel's tolerance is used, which means that only segmented curb pixels located at a distance of less than 0.1$m$ with respect to the curb labels are considered as true positive.
\begin{table}[!h]
	\caption{Binary segmentation results with different datasets}
	\label{tab:1}
	\centering
	\vspace{-0.8em}
	\begin{tabular}{cccccc}
		\toprule[2pt]
		Dataset  & Precision       & Recall          & F-1 score       & Image size & Tolerance \\ 
		\midrule[2pt]
		{\cite{Suleymanov2019}} & 0.8819          & 0.8921          & 0.8870          & 480*480    & 1 pixel   \\ 
		\specialrule{0em}{1pt}{1pt}
		\hline
		\specialrule{0em}{1pt}{1pt}
		{\cite{Younghwa2021}} & 0.9391          & 0.9427          & 0.9408          & 416*320    & 1 pixel   \\ 
		\specialrule{0em}{1pt}{1pt}
		\hline
		\specialrule{0em}{1pt}{1pt}
		Ours     & \textbf{0.9861} & \textbf{0.9785} & \textbf{0.9818} & 512*384    & 1 pixel   \\ \bottomrule[2pt]
	\end{tabular}
\end{table}


To ensure a fair comparison with other datasets, we conduct experiments only using sequence 00 (4,541 frames) of our curb dataset. $1/10$ frames in the sequence 00 are randomly selected as the valid set, and the rest are used as the training set. To train the network, the pixel-wise binary cross-entropy loss is employed as semantic loss function and the discriminative loss\cite{DeBrabandere2017} is used in our instance loss function.

In order to investigate the effect of partition methods splitting the training set and the valid set in curb binary segmentation, as shown in Table \ref{tab:2}, a comparative experiment is conducted on sequence 00 with different partition approaches, and the curb dilating kernel pixel size is set to 7. In the front-back partition approach, the first 9/10 frames in the sequence are selected as the training set and the rest as the valid set. Since the frames of valid set randomly selected in the sequence have higher similarity with the adjacent frames in the training set, the random partition approach performs better in this experiment.
\begin{table}[h]
	\centering
	\caption{Binary segmentation results with different partitions}  
	\label{tab:2}
		\vspace{-0.8em}
	\begin{tabular}{ccccc}
		\toprule[2pt]
		Partition & Precision       & Recall          & F-1 score       & Tolerance \\ 
		\midrule[2pt]
		Front-back    & 0.9243          & 0.9041          & 0.9140          & 1 pixel   \\ 
		\specialrule{0em}{1pt}{1pt}
		\hline
		\specialrule{0em}{1pt}{1pt}
		Random        & \textbf{0.9861} & \textbf{0.9785} & \textbf{0.9818} & 1 pixel   \\ 
		\bottomrule[2pt]
	\end{tabular}
\end{table}

Furthermore, the impact of different dilating kernel pixel sizes is analyzed and shown in Table \ref{tab:3}. The resolution of BEV representation is set to 0.1$m/pixel$ and the partition approach is set to random partition. A larger kernel size will result in more a favorable balance of positive and negative pixels, but greater position errors in binary segmentation results.

\begin{table}[h]
	\newcommand{\tabincell}[2]{\begin{tabular}{@{}#1@{}}#2\end{tabular}}  
	\centering
	\caption{Binary segmentation results with different kernel sizes}
	\label{tab:3}
	\vspace{-0.8em}
	\begin{tabular}{cccccc}
		\toprule[2pt]
		\tabincell{c}{Kernel \\ size}  & \tabincell{c}{Position \\ error}  & Precision & Recall & F-1 score & Tolerance \\ 
		\midrule[2pt]
		5           & $\bf{\pm 0.2m}$     & 0.9654    & 0.9539 & 0.9596    & 1 pixel   \\ 
		\specialrule{0em}{1pt}{1pt}
		\hline
		\specialrule{0em}{1pt}{1pt}
		7           & $\pm 0.3m$     & 0.9861    & 0.9785 & 0.9818    & 1 pixel   \\ 
		\specialrule{0em}{1pt}{1pt}
		\hline
		\specialrule{0em}{1pt}{1pt}
		9           & $\pm 0.4m$     & \textbf{0.9906}    & \textbf{0.9892} & \textbf{0.9899}    & 1 pixel   \\ 
		\bottomrule[2pt]
	\end{tabular}
\end{table}

\begin{table}[htb]
	\newcommand{\tabincell}[2]{\begin{tabular}{@{}#1@{}}#2\end{tabular}}  
	\centering
	\caption{Instance segmentation results}
	\label{tab:4}
	\vspace{-0.8em}
	\begin{tabular}{ccccc}
		\toprule[2pt]
		\tabincell{c}{IoU \\ threshold}  & Precision       & Recall          & F-1 score    & Tolerance   \\ 
		\midrule[2pt]
		0.5           & \textbf{0.9762}          & \textbf{0.9875}          & \textbf{0.9818}       & 1 pixel   \\ 
		\specialrule{0em}{1pt}{1pt}
		\hline
		\specialrule{0em}{1pt}{1pt}
		0.7           & 0.9533          & 0.9643          & 0.9588       & 1 pixel    \\ 
		\specialrule{0em}{1pt}{1pt}
		\bottomrule[2pt]
	\end{tabular}
\end{table}

Finally, the performance of curb instance segmentation is demonstrated in Table \ref{tab:4}. The partition approach is set to random partition and the curb dilating kernel pixel size is set to 7. True positive curb instances are with IoU values beyond the IoU threshold. The curb instance segmentation implemented on our dataset achieve promising results, and a higher IoU threshold contributes to a lower precision and recall.

\subsection{Semantic Segmentation with Curbs}
Semantic segmentation of LiDAR data is a point-wise classification task. Raw point clouds are projected into the dilated label image in Fig. \ref{fig:pic8}, and points that fall on the positive pixels are labeled as curb category. These points were previously classified as road, sidewalk or vegetation category, but now classified as curb category. The semantic segmentation network\cite{Gerdzhev2020} is employed again to train the data with curbs. Ground truths and predictions are shown in Fig. \ref{fig:pic10}. Following the original semantic segmentation challenge in \cite{Behley2019}, we train the network on sequences 00-07 and 09-10, and evaluate on sequence 08. As shown in Table \ref{tab:5}, segmentation results upon the dataset with curb category achieve comparable performances with the original dataset (20 categories), proving the validity of our annotations in the semantic segmentation task.

\begin{table}[htb]
	\centering
	\caption{Semantic segmentation results}
	\label{tab:5}
		\vspace{-0.8em}
	\begin{tabular}{cccccc}
		\toprule[2pt]
		Categories & mIoU  & road  & sidewalk  & vegetation & curb \\ 
		\midrule[2pt]
		20                & \textbf{0.611} &0.908 & 0.753    & \textbf{0.841}    &  $/$   \\ 
		\specialrule{0em}{1pt}{1pt}
		\hline
		\specialrule{0em}{1pt}{1pt}
		20+curb        & 0.601 & \textbf{0.935} & \textbf{0.785} & 0.814 & \textbf{0.709}    \\ 
		\bottomrule[2pt]
	\end{tabular}
\end{table}

\section{CONCLUSIONS}

In this paper, we presented an efficient two-stage curb labeling method which can generate point-wise and instance-wise curb annotations on LiDAR data. A range of baseline experiments of both instance segmentation and semantic segmentation have been implemented and evaluated on a generated dataset by our proposed method. The generated curb dataset has been released.

	\bibliographystyle{IEEEtran}
    \bibliography{icra.bib}

\end{document}